\documentclass[conference]{IEEEtran}
\IEEEoverridecommandlockouts

\usepackage{cite}
\usepackage{amsmath,amssymb,amsfonts}
\usepackage{algorithmic}
\usepackage{graphicx}
\usepackage{textcomp}
\usepackage{xcolor}
\usepackage{booktabs}
\usepackage{url}
\usepackage{hyperref}

\def\BibTeX{{\rm B\kern-.05em{\sc i\kern-.025em b}\kern-.08em
 T\kern-.1667em\lower.7ex\hbox{E}\kern-.125emX}}
\begin{document}

\title{DMAVA: Distributed Multi-Autonomous Vehicle Architecture Using Autoware}

\author{
\IEEEauthorblockN{Zubair Islam}
\IEEEauthorblockA{
\textit{Faculty of Engineering and Applied Science} \\
\textit{Ontario Tech University}\\
Oshawa, Canada \\
zubair.islam@ontariotechu.net}
\and
\IEEEauthorblockN{Mohamed El-Darieby}
\IEEEauthorblockA{
\textit{Faculty of Engineering and Applied Science} \\
\textit{Ontario Tech University}\\
Oshawa, Canada \\
Mohamed.El-Darieby@ontariotechu.ca}
}

\maketitle

\begin{abstract}
Simulating and validating coordination among multiple autonomous vehicles remains challenging, as many existing simulation architectures are limited to single-vehicle operation or rely on centralized control. This paper presents the Distributed Multi-Autonomous Vehicle Architecture (DMAVA), a simulation architecture that enables concurrent execution of multiple independent vehicle autonomy stacks distributed across multiple physical hosts within a shared simulation environment. Each vehicle operates its own complete autonomous driving stack while maintaining coordinated behavior through a data-centric communication layer. The proposed system integrates ROS 2 Humble, Autoware Universe, AWSIM Labs, and Zenoh to support high data accuracy and controllability during multi-vehicle simulation, enabling consistent perception, planning, and control behavior under distributed execution. Experiments conducted on multiple-host configurations demonstrate stable localization, reliable inter-host communication, and consistent closed-loop control under distributed execution. DMAVA also serves as a foundation for Multi-Vehicle Autonomous Valet Parking, demonstrating its extensibility toward higher-level cooperative autonomy. Demo videos and source code are available at: \href{https://github.com/zubxxr/distributed-multi-autonomous-vehicle-architecture}{\textbf{https://github.com/zubxxr/distributed-multi-autonomous-vehicle-architecture}}.

\end{abstract}


\section{Introduction}
\label{introduction}

Multi-vehicle autonomous driving (AD) simulations are essential for evaluating realistic interactions among autonomous vehicles (AVs), including physical dynamics, network-induced effects, and coordination strategies, as well as perception consistency, distributed decision-making under partial observability, and system scalability. Validation typically progresses through staged approaches such as model-in-the-loop \cite{b1}, software-in-the-loop (SIL), hardware-in-the-loop, and vehicle-in-the-loop (VIL) \cite{b2, b3}, in which simulation environments play a critical role in balancing accuracy, scalability, and cost.

Existing architectures present a fundamental trade-off between scalability and fidelity. Traffic simulators such as SUMO \cite{b4} and VISSIM \cite{b5} support large-scale multi-agent experimentation but operate at a macroscopic level and lack realistic sensor modeling required for full-stack AD evaluation. Commercial-grade simulators such as CarSim \cite{b6}, CarMaker \cite{b7}, and SCANeR \cite{b8} emphasize vehicle dynamics modeling and controller validation within SIL and VIL workflows. However, existing simulation-supported VIL-based studies primarily focus on single-vehicle configurations or controller-level validation \cite{b9, b10, b11}. As a result, these tools do not natively support distributed, ROS~2-based execution of multiple independent autonomy stacks within a shared simulation environment. Consequently, current validation pipelines struggle to simultaneously achieve realistic sensor fidelity, distributed execution, and coordinated multi-AV evaluation.

This paper presents the Distributed Multi-AV Architecture (DMAVA), which enables multiple concurrent Autoware \cite{b12} instances within a simulation environment using AWSIM Labs \cite{b13} and Zenoh \cite{b14}. DMAVA bridges SIL simulation and VIL-like data realism while supporting higher-level cooperative behaviors like distributed autonomous valet parking, demonstrated in our companion work \cite{b15}.

This paper is organized as follows. Section~\ref{related_works} reviews related work on current simulation architectures. Section~\ref{system_design} presents the system design and architecture of DMAVA, Section~\ref{experiments_results} describes the experimental setup and system performance, and Section~\ref{discussion_conclusions} concludes the paper with a discussion of strengths, limitations, and future directions.

\section{Related Works}
\label{related_works}
A wide range of simulators and visualization tools have been adopted for AD research and development. CARLA \cite{b16}, built on Unreal Engine, provides flexible scenario control and realistic rendering but requires additional middleware for ROS~2 integration. AWSIM \cite{b17} and AWSIM Labs, developed on the Unity Engine, are the official simulators for Autoware and share a common sensor and ROS~2 framework, though they are primarily designed for single-vehicle operation. Comparative analysis in \cite{b18} found AWSIM superior for LiDAR-based perception and Autoware integration, while CARLA remains stronger for end to end testing. Lightweight engines such as Godot \cite{b19} have also been explored within the Autoware ecosystem for visualization purposes, but existing implementations function as visualization frontends only and do not natively provide sensor generation or closed-loop simulation. Autoware has been demonstrated to interoperate with all of these simulation engines \cite{b20, b21, b17, b13}, enabling testing across different rendering pipelines and runtime environments. However, these environments are designed primarily for single-vehicle operation, limiting the ability to evaluate collaborative behaviors across multiple systems. For distributed multi-AV simulation, execution efficiency and runtime stability are critical. While AWSIM relies on the High-Definition Render Pipeline to achieve high-fidelity visualization, AWSIM Labs employs the Universal Render Pipeline to reduce rendering overhead and improve performance. Given its native compatibility with Autoware, reduced integration overhead compared to CARLA, and improved execution efficiency, AWSIM Labs was selected as the simulation platform in this work, making DMAVA better suited for scalable, distributed multi-AV execution.

Within the Autoware ecosystem, the only publicly documented example of multi-AV operation is a demonstration by ADLINK Technology \cite{b22}, consisting of two Autoware instances controlling two vehicles in CARLA via Zenoh. All components were executed on a single host, with CARLA rendered off-screen and instead visualized through Autoware’s RViz camera panel. This design limited scalability and visualization fidelity and resulted in observable rendering lag and resource contention during execution. This work extends this concept by implementing a multi-host configuration within AWSIM Labs rather than CARLA. The containerized deployment used in the ADLINK experiment informed the design of DMAVA, which distributes multiple containers across multiple physical hosts to improve scalability and maintain consistent, synchronized operation under distributed execution. By distributing simulation and autonomy workloads, AWSIM Labs can be rendered normally without requiring off-screen rendering, avoiding the visualization and resource contention constraints observed in single-host setups.

Beyond the Autoware domain, distributed multi-AV simulation architectures such as Sky-Drive \cite{b23} have explored large-scale collaborative environments. Sky-Drive extends CARLA to support synchronized multi-AV simulation across multiple hosts with hardware-in-the-loop and human-in-the-loop capability. While it emphasizes human–AI collaboration and remote experimentation, its design priorities differ from Autoware-centric distributed simulation frameworks.

Collectively, these studies highlight the growing importance of distributed simulation for evaluating  multi-AV systems. DMAVA builds upon these foundations by introducing a fully integrated, ROS 2-native, Zenoh-enabled multi-AV simulation environment within AWSIM Labs that achieves high data accuracy and controllability under distributed operation, while maintaining scalability and acceptable experimental cost for multi-AV validation in the Autoware ecosystem.

\section{System Design and Architecture}
\label{system_design}

This section describes the system design and architectural principles of the DMAVA. 

\subsection{DMAVA Structure and Workflow Organization}
DMAVA integrates Autoware Universe, AWSIM Labs, and Zenoh within a distributed simulation pipeline. AWSIM Labs is responsible for vehicle physics, motion, and sensor generation, while Autoware operates as an external AD controller that consumes simulated sensor data and publishes control commands. Zenoh enables inter-host communication between distributed components, allowing multiple AV stacks to operate concurrently across separate physical hosts.

The system is organized into five functional workflows that define the overall data flow and module interaction across hosts, illustrated in Figure~\ref{multi_vehicle_architecture}, and detailed below:

\begin{figure*}[t]
\centering
\includegraphics[width=0.7\linewidth]{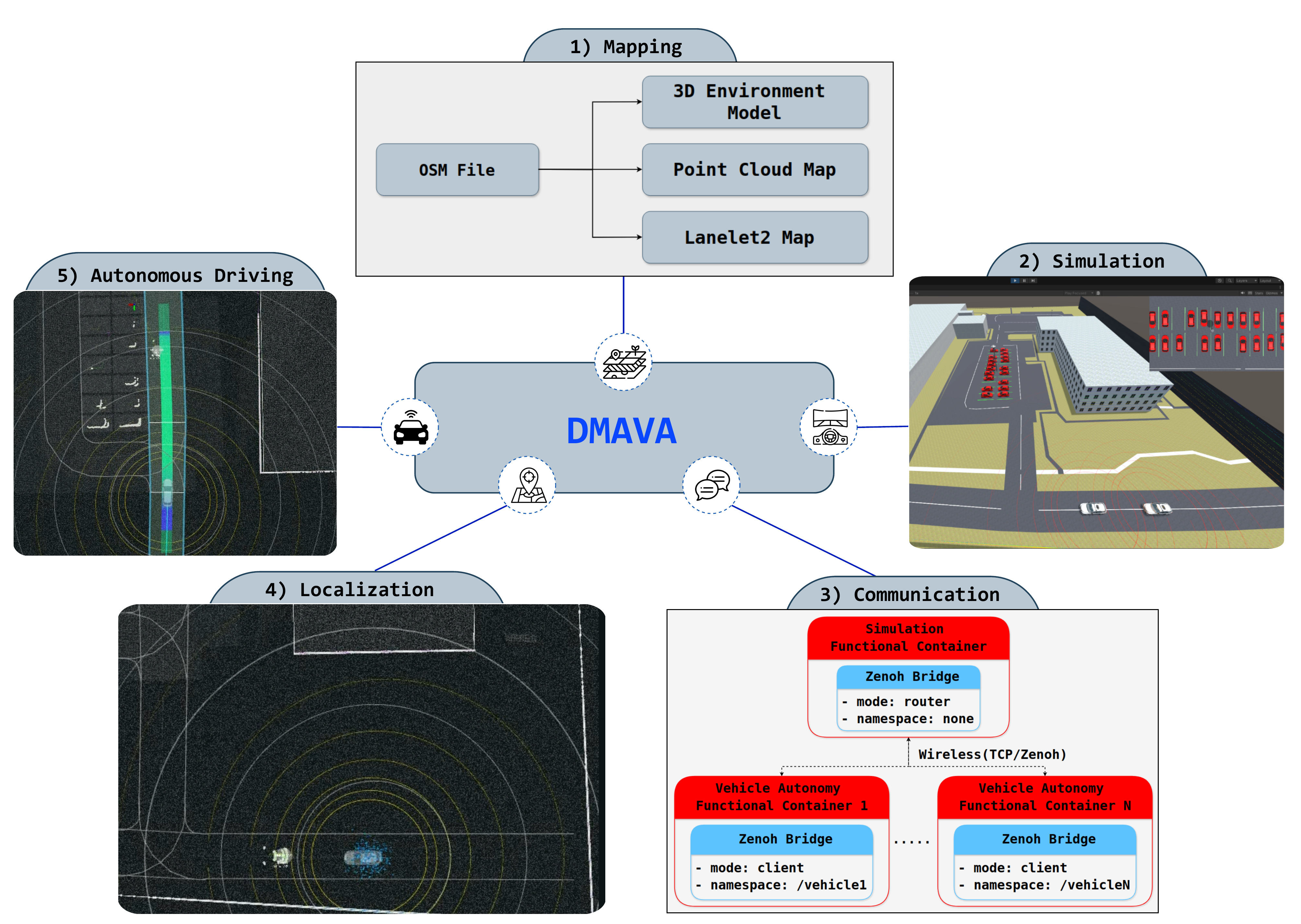}
\caption{Proposed DMAVA showing the five-workflow design.}
\label{multi_vehicle_architecture}
\end{figure*}

\begin{enumerate}
 \item \textit{Mapping Workflow: }Converts OpenStreetMap regions into 3D meshes, point-cloud maps, and Lanelet2 vector files using the AV Map Creation Workflow \cite{b24}.
 \item \textit{Simulation Workflow: }Loads these assets in AWSIM Labs, configures multiple ego vehicles equipped with virtual LiDAR, cameras, and IMU sensors, and assigns distinct ROS 2 namespaces to each vehicle to prevent topic collisions.
 \item \textit{Communication Workflow:} Employs Zenoh to exchange ROS 2 topics between hosts with namespace-aware routing.
 \item \textit{Localization Workflow:} Aligns LiDAR scans from AWSIM Labs with the point cloud map in Autoware.
 \item \textit{Autonomous-Driving Workflow: }Executes perception, planning, and control within Autoware to achieve closed-loop behavior.
\end{enumerate}

Each workflow builds upon the previous one, forming a modular pipeline that can be scaled or reconfigured for different experimental scenarios. These workflows represent functional responsibilities and data flow within the system and are encapsulated as containerized execution units, remaining independent of how they are deployed or co-located across physical hosts. Communication is provided through embedded Zenoh middleware rather than as a standalone containerized workflow. Related workflow containers are grouped into higher-level functional containers based on their execution role.

\begin{figure*}[t]
 \centering
 \includegraphics[width=\linewidth]{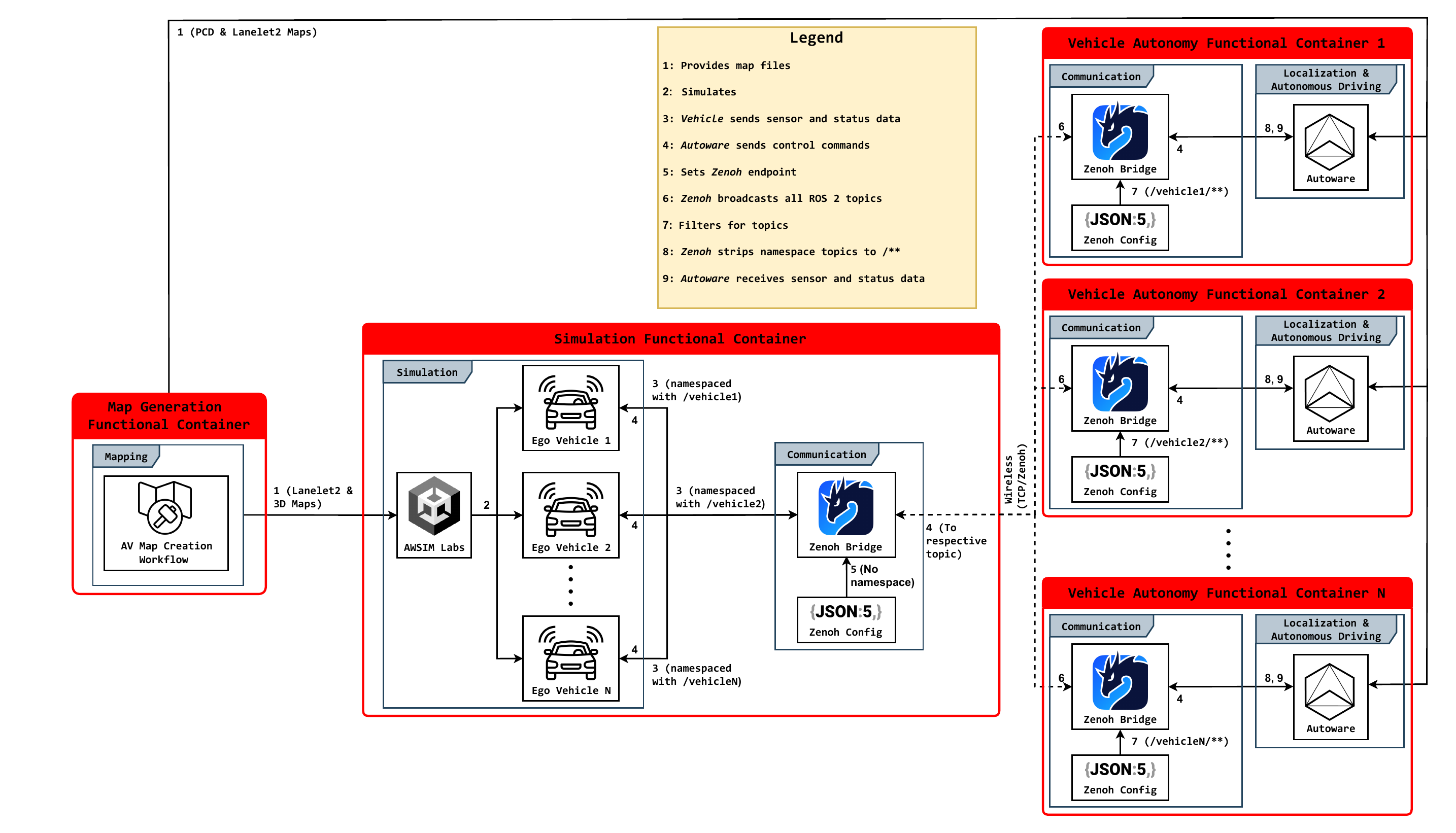}
 \caption{System-level container-based architecture of DMAVA.}
 \label{system_workflow}
\end{figure*}

Figure~\ref{system_workflow} illustrates the container-based design of DMAVA, organized into three functional containers: Map Generation, Simulation, and Vehicle Autonomy. The Map Generation functional container is executed offline on any available host prior to simulation and is responsible for producing the Lanelet2 and point cloud map assets required by the system. These map artifacts are then provided to the Simulation functional container for environment initialization and to the Vehicle Autonomy functional containers for localization and planning. Within the Simulation functional container, AWSIM Labs loads the generated environment and simulates multiple ego vehicles within a shared virtual scene. Each simulated vehicle is equipped with a configurable sensor suite and publishes sensor and vehicle status data under a unique ROS 2 namespace.

Each Vehicle Autonomy functional container runs an independent Autoware instance responsible for per-vehicle localization, planning, and control. All functional containers, excluding Map Generation, embed Zenoh which selectively exposes and filters ROS 2 topics, enabling sensor data and control commands to be exchanged between the simulation and autonomy containers independent of physical host placement. This design preserves per-vehicle isolation, as vehicles do not exchange direct V2V messages and each Autoware instance communicates exclusively with the simulator through Zenoh.

\subsection{Zenoh Communication Design}
\label{zenoh}

Zenoh serves as the communication layer using a lightweight, data-centric publish–subscribe protocol to selectively exchange ROS~2 topics across distributed containers. Its behavior is configured using JSON5 configuration files that define communication parameters, namespace handling rules, and topic filtering, allowing fine-grained control over how data is transmitted and received. Zenoh transports sensor data and control messages between AWSIM Labs and Autoware instances while preserving per-vehicle isolation. To support scalable multi-vehicle execution, each vehicle autonomy instance operates within a unique ROS~2 namespace. Namespaces are used to prevent topic and service collisions when multiple vehicles are simulated concurrently and to enable isolation between independent autonomy stacks.

The Zenoh design follows a container-level abstraction in which the Simulation functional container runs a Zenoh instance configured in router mode without a namespace. This router publishes ROS~2 topics generated by AWSIM Labs, which are already namespaced per simulated vehicle. Each Vehicle Autonomy functional container runs a Zenoh instance configured in client mode and is assigned a specific vehicle namespace, such as \texttt{/vehicle1}, \texttt{/vehicle2}, up to \texttt{/vehicleN}, as depicted in Figure~\ref{system_workflow}. Each client selectively subscribes to topics within its assigned namespace and strips the namespace prefix before delivering messages to the local Autoware instance, which operates without namespaces. After processing, Autoware publishes control commands without namespace prefixes, and the local Zenoh client re-applies the corresponding vehicle namespace before forwarding these commands back to the Simulation functional container. This bidirectional namespace translation enables compatibility with Autoware’s default topic structure while preserving per-vehicle isolation within the shared simulation environment. Additional vehicle autonomy containers can be integrated by assigning a new namespace and registering a corresponding Zenoh client, without modifying the existing communication configuration.

To ensure stable runtime performance and limit unnecessary network overhead, high-bandwidth image topics and parameter-related services are explicitly filtered in the configuration file. Excluding these topics reduces cross-container traffic and prevents large data streams from impacting localization and control performance. The corresponding JSON5 configuration files are provided in \cite{b25}.

\subsection{Refinement of Localization Robustness}
\label{autoware_localization}

Autoware serves as the AD software that is built up of perception, localization, planning, and control stacks, all built on ROS 2 Humble. While all the stacks operated successfully, the localization stack required refinement to improve robustness in simulation. Autoware uses the Normal Distributions Transform (NDT) algorithm to align incoming LiDAR scans from AWSIM Labs with the point cloud map. However, early testing revealed instability with localization near the vehicle spawn area, where the NDT-based scan matching frequently produced inconsistent pose estimates due to limited geometric features in the environment. To address this, the map was refined in Blender \cite{b26} by trimming unused geometry and adding planar walls to provide stronger scan-matching surfaces. The updated mesh was then reinserted into the Mapping Workflow, which was enhanced to incorporate the manual mesh geometry addition step, marking a refinement of the workflow for more accurate localization.

Further refinement of the localization parameters in Autoware was needed to balance accuracy with computational load. Host 1 used aggressive parameter settings to prioritize precision, while Host 2 employed lighter configurations to align with its hardware limitations. The complete Autoware localization configurations are provided in \cite{b25}.

\subsection{Extending AWSIM Labs for Multi-Vehicle Simulation}

By default, AWSIM Labs supports only a single active vehicle. This limitation arises from its reliance on the Vehicle Physics Pro (VPP) Unity asset, which, under the free community license, restricts simulation to a single vehicle per scene. Enabling multiple vehicles within AWSIM Labs therefore requires a paid professional or enterprise VPP license, making native multi-AV simulation infeasible without modification \cite{b27}. To overcome this fundamental constraint, AWSIM Labs was restructured to bypass the VPP dependency while preserving full compatibility with existing ROS 2 interfaces and Autoware integration. Specifically, the original AWSIM simulator, which does not rely on VPP and supports the simulation of multiple vehicles, was used as a reference. However, the vehicles in AWSIM do not operate as independent autonomous agents and lack dedicated autonomy stacks.

The AWSIM vehicle prefab was duplicated, adapted, and imported into AWSIM Labs to enable multiple vehicles to coexist within the same simulation scene. Each vehicle instance was configured with an independent ROS~2 namespace and clock synchronization, ensuring isolated topic spaces and preventing inter-vehicle interference. This extension transforms AWSIM Labs from a single-vehicle simulator into a platform capable of hosting multiple concurrently operating AVs with independent full-stack autonomy.

\section{Experiments and Results}
\label{experiments_results}

DMAVA was evaluated under two and three host distributed configurations to assess localization stability, Zenoh communication reliability, system scalability, communication latency, and resource utilization.

\subsection{Experimental Setup}
DMAVA was deployed under two distributed configurations: a two-host setup and a three-host extension. Initial single-host testing revealed significant resource contention and unstable execution when multiple Autoware instances were launched concurrently, making sustained operation impractical. 

Although DMAVA supports fully distributed execution, resource constraints influenced container placement during experimental evaluation. Table~\ref{tab:hardware_specs} lists the hardware platforms used, with the Nitro PC serving as the primary simulation host. In the two-host configuration, the Simulation functional container and one Vehicle Autonomy functional container were co-located on the Nitro PC, while a second Vehicle Autonomy functional container was deployed on the ROG Laptop. This configuration served as the primary setup for validating stable operation. The three-host configuration extended this deployment by assigning an additional Vehicle Autonomy functional container to the Victus Laptop, enabling evaluation of scalability and performance under increased computational load. This co-location was performed solely for practical resource management and does not represent a limitation of the proposed architecture. Logically, all system components remain independent and may be deployed on separate physical hosts if sufficient computational resources are available. All experiments were conducted using Autoware Universe (release/2024.11 \cite{b28}), AWSIM Labs (main branch, modified), and the Zenoh bridge for ROS 2 (release/1.7.2 \cite{b29}), running on Ubuntu 22.04 LTS with ROS 2 Humble Hawksbill. 

\begin{table}[t]
\centering
\caption{Hardware Specifications of Host Machines}
\label{tab:hardware_specs}
\begin{tabular}{|c|c|c|c|}
\hline
\textbf{Host} & \textbf{CPU} & \textbf{GPU} & \textbf{RAM} \\
\hline
(1) Nitro PC & i7-12700F & RTX 3060 & 24 GB \\
(2) ROG Laptop & Ryzen 7 4800HS & RTX 2060 Max-Q & 24 GB \\
(3) Victus Laptop & i5-12500H & RTX 4050 & 16 GB \\
\hline
\end{tabular}

\vspace{4pt}
\footnotesize{All hosts were running NVIDIA driver version 575.}
\end{table}

Initial testing was conducted with the ROG Laptop serving as the simulation host. This configuration revealed rendering lag and synchronization instability due to limited GPU and memory resources. As a result, the simulation workload was migrated to the Nitro PC, which served as the primary simulation host in all experiments. This configuration achieved stable rendering, consistent synchronization, and reliable runtime performance. System startup followed a defined sequence to ensure consistent synchronization across all hosts and reliable topic alignment. Host 1 initialized the simulation by launching AWSIM Labs, followed by its Autoware instance. Host 2 then launched its own Autoware instance. Once both Autoware stacks were active, Zenoh was executed on each host to establish inter-host communication. This startup order ensured consistent initialization, synchronized topic discovery, and stable message exchange during runtime.

\subsection{Localization and Zenoh Performance}
Each Autoware instance performed localization using the NDT Scan Matcher with the generated point cloud map. Using the refined map and tuned NDT parameters (Section~\ref{autoware_localization}), localization was evaluated across 24 startup trials conducted under identical conditions. Automatic convergence was achieved in 22 trials (92\%) without manual intervention. In the remaining trials, convergence did not occur during initial startup and was resolved by manually providing an initial pose estimate near the expected vehicle location, after which the localization stack converged successfully. Once converged, localization remained stable throughout operation.

Zenoh behavior was further evaluated by monitoring connection stability, startup consistency, and cross-host data delivery. Zenoh instances on all hosts operated in router mode, resulting in frequent startup failures and unstable data delivery, with vehicles failing to localize in Autoware in approximately 9 out of 10 trials. Reconfiguring the Vehicle Autonomy hosts to operate in client mode, while keeping the Simulation host in router mode, eliminated these failures, resulting in reliable localization initialization and consistent cross-host synchronization across all trials.

Additional stress testing under increased  loads revealed further communication bottlenecks. During three-host operation, large image topics generated excessive bandwidth demand, delaying message delivery and causing intermittent desynchronization, while parameter-related services produced inconsistent responses during cross-host service calls. To mitigate these effects, both image topics and parameter-related services were explicitly filtered in Zenoh’s configuration. After applying these filters, both two- and three-host configurations maintained stable  data exchange for multi-AV operations.

\subsection{Distributed Validation and Scalability Testing}
\label{distributed_validation_scalability_testing}

\begin{figure}[t]
 \centering
 \includegraphics[width=\linewidth]{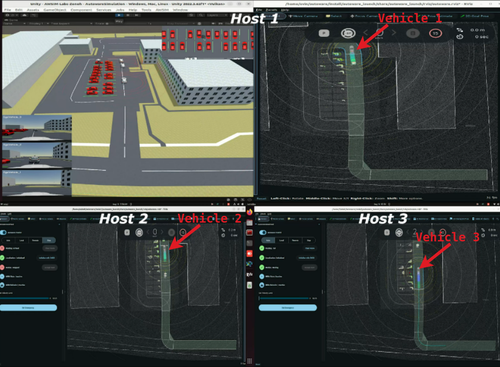}
 \caption{Consolidated visualization of two-host and three-host validation, demonstrating synchronized topic exchange and namespace isolation across distributed Autoware systems.}
 \label{multi_host}
\end{figure}

DMAVA demonstrated stable multi-vehicle operation in the two-host configuration, where each vehicle operated under its own Autoware instance with synchronized localization and topic exchange. Both vehicles received goals, followed planned trajectories, and completed navigation tasks autonomously, demonstrating consistent motion, accurate path following, and stable planner behavior. When extending the system from two to three vehicles, autonomous operation was maintained across all hosts. Each vehicle continued to receive goals, generate valid trajectories, and execute closed loop planning and control without manual intervention. Results for both configurations are shown in the consolidated visualization in Figure~\ref{multi_host}.

However, intermittent instability was observed during prolonged three-host execution. Although these effects did not prevent task completion, they motivated further investigation into system behavior under increased load. System performance and communication timing are examined in the following section to identify contributing factors. Video demonstrations of both the two-host and three-host configurations are provided in \cite{b25}.

\subsection{System Performance and Communication Timing}
\label{sec:communication_latency}

To characterize system behavior under increased load, memory availability, CPU utilization, and inter-host communication latency were monitored for both two-host and three-host configurations. Across all configurations, available memory remained high and no swapping occurred (Table~\ref{tab:memory_after_launch}), indicating that memory was not the primary limiting factor contributing to the observed intermittent instability.

\begin{table}[t]
\centering
\caption{Available System Memory for Two- and Three-Host Configurations}
\label{tab:memory_after_launch}
\begin{tabular}{|c|c|c|}
\hline
\textbf{Setup} & \textbf{Host} & \textbf{Available Memory} \\
\hline
Two-Host & ROG Laptop & 11 GiB \\
 & Nitro PC & 9.8 GiB \\
\hline
Three-Host & ROG Laptop & 8.0 GiB \\
 & Nitro PC & 8.3 GiB \\
 & Victus Laptop & 5.4 GiB \\
\hline
\end{tabular}
\end{table}

CPU utilization was recorded over the full system lifecycle, beginning after successful localization of all vehicles, and continuing through goal assignment and autonomous execution. CPU usage remained within stable operating limits, with only transient peaks during planning and control, indicating that CPU was not the primary contributor to the observed intermittent instability (Table~\ref{tab:cpu_utilization}).

\begin{table}[t]
\centering
\caption{CPU Utilization for Two-Host and Three-Host Configurations}
\label{tab:cpu_utilization}
\begin{tabular}{|c|c|c|c|}
\hline
\textbf{Configuration} & \textbf{Host} & \textbf{Mean CPU (\%)} & \textbf{Peak CPU (\%)} \\
\hline
Two-Host & ROG Laptop & 57 & $<90$ \\
 & Nitro PC & 44 & $<50$ \\
\hline
Three-Host & ROG Laptop & 67 & $<90$ \\
 & Nitro PC & 50 & $<55$ \\
 & Victus Laptop & 42 & $<60$ \\
\hline
\end{tabular}
\end{table}

Inter-host communication latency was evaluated using round-trip time (RTT) measurements collected via ROS 2 probe nodes bridged through Zenoh. The Nitro PC timestamped outgoing probe messages, which were echoed by secondary hosts, enabling end-to-end latency measurement using a single time reference. Measurements were collected at 10 Hz under three operating conditions: baseline (Zenoh and probes only), system idle (AWSIM and Autoware initialized without vehicle motion), and active operation during AD.

Initial experiments were conducted in a two-host (Nitro, ROG) configuration using an existing household network, where the residential router was located on the floor below and across the house from the experiment hosts. The Nitro PC was connected wirelessly, while the ROG Laptop was connected via a wired Ethernet link. RTT results for the baseline, idle, and active phases are summarized in Table~\ref{tab:rtt_two_host}, where RTT is reported as mean ± standard deviation, with the standard deviation representing latency jitter. Mean RTT remained below 6.5 ms across all phases, with increased jitter observed during idle and active operation. No packet loss was observed, and RTT variability did not exceed the reported standard deviation values.

\begin{table}[t]
\centering
\caption{RTT Measurements Under Two-Host Operation Using Existing Household Network Infrastructure}
\label{tab:rtt_two_host}
\begin{tabular}{|c|c|c|c|}
\hline
\textbf{Phase} & \textbf{RTT (ms)} & \textbf{Max RTT (ms)} & \textbf{Samples} \\
\hline
Baseline & $3.78 \pm 1.67$ & 15.57 & 631 \\
Idle system & $6.10 \pm 5.70$ & 73.02 & 604 \\
Active operation& $3.85 \pm 4.55$ & 52.42 & 1135 \\
\hline
\end{tabular}
\end{table}

Extending this configuration to three hosts resulted in frequent packet loss and intermittent communication failures during active operation, preventing reliable execution. As a result, the network setup was revised to use a Wi-Fi 6 (802.11ax) dual-band access point (TP-Link Archer AX23), with all hosts connected over short-range wireless links in the same physical location. Under the revised network configuration, RTT measurements were collected during active operation for both two-host and three-host setups (Table~\ref{tab:rtt_three_host}).

\begin{table}[t]
\centering
\caption{RTT Measurements During Three-Host Active Operation Under Dedicated Local Access Point}
\label{tab:rtt_three_host}
\begin{tabular}{|c|c|c|c|}
\hline
\textbf{Configuration} & \textbf{RTT (ms)} & \textbf{Max RTT (ms)} & \textbf{Samples} \\
\hline
Two-Host & $41.03 \pm 61.16$ & 352.49 & 1134 \\
\hline
Three-Host (ROG) & $61.26 \pm 87.95$ & 625.10 & 1430 \\
\hline
Three-Host (Victus) & $32.5 \pm 44.64$ & 309.41 & 1481 \\
\hline
\end{tabular}
\end{table}

While mean RTT and jitter increased compared to the original two-host configuration, RTT variation remained within the reported mean and standard deviation ranges, and no sustained packet loss was observed, indicating that nominal RTT magnitude alone does not explain the observed instability under multi-host operation. These results indicate that DMAVA operates as a distributed architecture in which network stability and contention characteristics are more critical to reliable multi-host operation than nominal message latency alone.

\section{Discussion and Conclusions}
\label{discussion_conclusions}
This work introduces DMAVA, which represents, to the authors’ knowledge, the first distributed multi-AV simulation architecture that integrates Autoware Universe, AWSIM Labs, and Zenoh within a unified and fully open-source environment. It coordinates multiple independent Autoware stacks across hosts without modifying Autoware internals using namespace isolation and selective topic exchange, and provides a scalable foundation for multi-vehicle research.

DMAVA is intentionally designed as a communication and execution backbone rather than a decision-making framework. Cooperative behaviors such as vehicle-to-vehicle negotiation, shared planning, or heterogeneous agent coordination are not embedded within the core system, ensuring that DMAVA remains a stable and reproducible infrastructure-level baseline. This design enables higher-level coordination logic to be layered on top of the architecture without modifying its core components, as demonstrated by the Distributed Multi-Vehicle Autonomous Valet Parking system \cite{b15}. 

Scaling the system from two to three hosts validated the scalability of the architecture while also revealing system-level constraints. Although intermittent instability was observed during prolonged three-host operation, experimental measurements confirmed that all hosts maintained sufficient CPU and memory headroom, and no sustained computational bottlenecks were identified. RTT analysis further showed that nominal latency magnitude alone did not explain the observed instability. Instead, system behavior correlated more strongly with communication stability and timing variability under concurrent multi-host operation. These results indicate that scalability in the evaluated setup was constrained primarily by networking and transport characteristics rather than by per-host computational capacity.

Future work will focus on improving robustness, deployment flexibility, and real world applicability. Cloud or edge offloading could reduce per host computational load, while extending Zenoh to handle high bandwidth image topics would enable distributed vision based perception. Deployment on embedded robotic platforms will facilitate validation under real world network conditions. In addition, future evaluations could leverage more homogeneous compute platforms and dedicated networking infrastructure to further reduce timing variability and improve stability under higher vehicle counts. To support reproducibility and continued extension, the DMAVA implementation, including configuration files and deployment instructions, is made publicly available as an open source reference that enables replication of the reported experiments and adaptation to additional vehicles or application domains. Collectively, these directions position DMAVA as a scalable foundation that can be extended beyond simulation toward more operational distributed multi AV deployments.

\vspace{12pt}

\end{document}